\documentclass{article}

\usepackage{times}

\usepackage{amsmath}
\usepackage{amssymb}
\usepackage{amsfonts,amsthm}

\usepackage{graphicx} 
\usepackage{subfigure} 

\usepackage{natbib}

\usepackage{algorithm}
\usepackage{algorithmic}

\usepackage{hyperref}

\newcommand{\E}{\mathbb{E}}
\newcommand{\R}{\mathbb{R}}
\DeclareMathOperator*{\argmax}{\arg\!\max}


\usepackage[accepted]{icml2014}

\icmltitlerunning{Marginal Structured SVM with Hidden Variables}

\begin{document} 

\twocolumn[
\icmltitle{Marginal Structured SVM with Hidden Variables}

\icmlauthor{Wei Ping}{wping@ics.uci.edu}
\icmlauthor{Qiang Liu}{qliu1@ics.uci.edu}
\icmlauthor{Alexander Ihler}{ihler@ics.uci.edu}
\icmladdress{Department of Computer Science, University of California, Irvine}

\icmlkeywords{Latent Variable Model, Discriminative, Structured Learning, Max-Margin}

\vskip 0.3in
]

\begin{abstract}
In this work, we propose the marginal structured SVM (MSSVM) for structured prediction with hidden variables. MSSVM properly accounts for the uncertainty of hidden variables, and can significantly outperform the previously proposed latent structured SVM~(LSSVM; \citet{Joachims09}) and other state-of-art methods,  especially when that uncertainty is large. 
Our method also results in a smoother objective function, making gradient-based optimization of MSSVMs converge significantly faster than for LSSVMs. 
We also show that our method consistently outperforms hidden conditional random fields~(HCRFs; \citet{Quattoni07}) on both simulated and real-world datasets. 
Furthermore, we propose a unified framework that includes both our and several other existing methods as special cases, and provides insights into the comparison of different models in practice. 
\end{abstract}

\section{Introduction}
Conditional random fields (CRFs) \cite{Lafferty01} and structured SVMs (SSVMs) \cite{Taskar03, Tsochantaridis05} are standard tools for structured prediction in many important domains, such as computer vision \cite{Nowozin11}, natural language processing \cite{Taskar07} and computational biology \citep[e.g.,][]{Li07, Sato05}. 
However, many practical cases are not well handled by these tools, due to the presence of latent variables or partially labeled datasets.
For example, one approach to image segmentation classifies each pixel into a predefined semantic category. While it is expensive to collect labels for every single pixel (perhaps even impossible for ambiguous regions), partially labeled data are relatively easy to obtain~\citep[e.g.,][]{Verbeek06}.
Examples also arise in natural language processing, such as semantic role labeling, where the semantic predictions are inherently coupled with latent syntactic relations \citep{Naradowsky12}. However, accurate syntactic annotations are unavailable in many language resources.

In past few years, several solutions have been proposed to address hidden variable problems in structured prediction. 
Perhaps the most notable of these are hidden conditional random fields (HCRFs) \cite{Quattoni07} and latent structured SVMs (LSSVMs) \cite{Joachims09}, which are derived from conditional random fields and structured SVMs, respectively.
However, both approaches have several shortcomings.
CRF-based models often perform worse than SSVM-based methods in practical datasets, especially when the number of training instances is small or the model assumptions are heavily violated \citep[e.g.,][]{Taskar03}. 
On the other hand, LSSVM relies on a joint maximum {\it a posteriori} (MAP) procedure that assigns the hidden variables to deterministic values, and does not take into account their uncertainty. 
Unfortunately, this can produce poor predictions of the output variables even for exact models \citep{Liu13}. 
A better approach is to average over possible states, corresponding to a \emph{marginal MAP} inference task \citep{Koller09, Liu13} that marginalizes the hidden variables before optimizing over the output variables.

{\bf Contributions.} 
We propose a novel structured SVM algorithm that takes into account the uncertainty of the hidden variables, by incorporating marginal MAP inference that  ``averages" over the possible hidden states. 
We show that our method performs significantly better than LSSVM and other state of art methods, especially when the uncertainty of the hidden variables is high. 
Our method also inherits the general advantages of structured SVMs and consistently outperforms HCRFs, especially when the training sample size is small. 
We also study the effect of different training algorithms under various models. In particular we show that gradient-based algorithms for our framework are much more efficient than for LSSVM, because our objective function is smoother than that of LSSVM as it marginalizes, instead of maximizes, over the hidden variables. 
Finally, we propose a unified framework that includes both our and existing methods as special cases, and provide general insights on the choice of models and optimization algorithms for practitioners. 

We organize the rest of the paper as follows. In Section~\ref{sec:relatedwork}, we introduce related work. We present background and notation in Section~\ref{sec:notations}, and derive our marginal structured SVM in Section~\ref{sec:MSSVM}. The unified framework is proposed in Section~\ref{sec:Unified Framework}.  Learning and inference algorithms for the model are presented in Section~\ref{sec:learning}. We report experimental results in Section~\ref{sec:experiments} and conclude the paper in Section~\ref{sec:conclusion}.

\section{Related Work}
\label{sec:relatedwork}
HCRFs naturally extend CRFs to include hidden variables, and have found numerous applications in areas such as object recognition \cite{Quattoni04} and gesture recognition \citep{Wang06}. 
HCRFs have the same pros and cons as general CRFs; in particular, they perform well when the model assumptions hold and when there are enough training instances, but may otherwise perform badly.
Alternatively, the LSSVM \citep{Joachims09} is an extension of structured SVM that handles hidden variables, with wide application in areas like object detection \cite{LongZhu10},  human action recognition \cite{Wang09}, document-level sentiment classification \cite{Yessen10} and link prediction \cite{Xu13}.
However, LSSVM relies on a joint MAP procedure, and may not perform well when a non-trivial uncertainty exists in the hidden variables. 
Recently, \citet{Schwing12} proposed an $\epsilon$-extension framework for discriminative graphical models with hidden variables that includes both HCRFs and LSSVM as special cases. 

A few recent works also incorporate uncertainty over hidden variables explicitly into their optimization
frameworks.
For example, \citet{Miller12} proposed a max margin min-entropy (M3E) model that 
minimizes an uncertainty measure on hidden variables while performing max-margin learning. 
They assume that minimizing hidden uncertainty will improve the output accuracy. This is valid in some applications, such as object detection, where reducing the uncertainty of object location can improve the category prediction. However, in cases like image segmentation, the missing labels may come from ambiguous regions, and maintaining that ambiguity can be important. 
In another work, \citet{Kumar12} proposes a learning procedure that encourages agreement between two separate models -- one for predicting outputs and another for representing the uncertainty over the hidden variables.  They model the uncertainty of hidden variable during training, and rely on a joint MAP procedure during prediction.

Our proposed method builds on recent work for marginal MAP inference \citep{Koller09, Liu13}, which averages over the hidden variables (or variables that are not of direct interest), and then optimizes over the output variables (or variables of direct interest).
In many domains, marginal MAP can provide significant improvement over joint MAP estimation, which jointly optimizes hidden and output variables; recent examples include blind deconvolution in computer vision \citep{Fergus06, Levin11} and relation extraction and semantic role labeling in natural language processing \citep{Naradowsky12}. 
Unfortunately, marginal MAP tasks on graphical models are notoriously difficult; marginal MAP can be NP-hard even when the underlying graphical model is tree-structured \citep{Koller09}. 
Recently, \citet{Liu13} proposed efficient variational algorithms that approximately solve marginal MAP. In our work, we use their mixed-product belief propagation algorithm as our inference component.

Sub-gradient decent (SGD) \cite{Ratliff07} and the concave-convex procedure (CCCP)\cite{Yuille03} are two popular training algorithms for structured prediction problems.
Generally, 
SGD is straightforward to implement and effective in practice, but may be slow to converge, especially on non-convex and non-smooth objective functions as arise in LSSVMs.
CCCP is a general framework for minimizing non-convex functions by transforming the non-convex optimization into a sequence of convex optimizations by iteratively linearizing the non-convex component of the objective.  
It has been applied widely in many areas of machine learning, particularly when hidden variables or missing data are involved. 
We explore both these training methods and compare them across the various models we consider. 

\section{Structured Prediction with Hidden Variables}
\label{sec:notations}
In this section we review the background on structured prediction with hidden variables. Assume we have structured input-output pairs $(x, y) \in \mathcal{X} \times \mathcal{Y}$, where $\mathcal{X}$, $\mathcal{Y}$ are the spaces of the input and output variables. In many applications, this input-output relationship is not only characterized by $(x, y)$, but also depends on some unobserved hidden or latent variables $h \in \mathcal{H}$. Suppose $(x, y, h)$ follows a conditional model,
\begin{align}
p(y, h|x; w) = \frac{1}{Z(x; w)} \exp \ [w^{T}\phi(x, y, h)], \label{e1}
\end{align}
where $\phi(x, y, h): \mathcal{X} \times \mathcal{Y} \times \mathcal{H} \rightarrow \mathbb{R}^D$ is a set of features which describe the relationships among the $(x, y, h)$, and $w \in \mathbb{R}^D$ are the corresponding  weights, or model parameters. The function $Z(x; w)$ is the normalization constant, or \emph{partition function},
\begin{equation}
Z(x; w) = \sum_{y, h} \exp \ [w^{T}\phi(x, y, h)]. \nonumber
\end{equation}
Assuming the weights $w$ are known, the LSSVM of \citet{Joachims09} decodes the output variables $y$ given input variables $x$ by performing a joint maximum \emph{a posteriori} (MAP) inference,
\begin{align}
 [ \tilde{y}(w), \tilde{h}(w) ]  &= \argmax_{(y, h) \in  \mathcal{Y} \times \mathcal{H} } p(y, h|x) \nonumber \\
&= \argmax_{(y, h) \in  \mathcal{Y} \times \mathcal{H}} \ w^{T}\phi(x, y, h).  \nonumber
\end{align}
This gives the optimal prediction of the ($y, h$)-pair, and one obtains a prediction on $y$ by simply discarding the $h$ component. 
Unfortunately, 
the optimal prediction for ($y, h$) jointly does not necessarily give an optimal prediction on $y$; instead, it may introduce strong biases even for simple cases (e.g., see Example 1 in \citet{Liu13}). Intuitively, the joint MAP prediction is ``overly optimistic'', since it deterministically assign the hidden variables to their most likely states; this approach is not robust to the inherent uncertainty in $h$, which may cause problems if that uncertainty is significant. 

To address this issue, we use marginal MAP predictor, 
\begin{eqnarray}
\hat{y}(w) &=& \argmax_{y\in \mathcal{Y}} \sum_{h} p(y,h|x; w) \nonumber\\
&=& \argmax_{y\in \mathcal{Y}} \log\sum_{h } \exp \ [w^{T}\phi(x, y, h)],  \label{e3}
\end{eqnarray}
which explicitly takes into account the uncertainty of the hidden variables. It should be noted that $\hat{y}(w)$ is in fact the Bayes optimal prediction of $y$, measured by zero-one loss.
The main contribution of this work is to introduce a novel structured SVM-based method for training the marginal MAP predictor, which significantly improves over previous methods.

\section{Marginal Structured SVM}
\label{sec:MSSVM}
In this section we derive our main method, the marginal structured SVM (MSSVM), which minimizes an upper bound of the empirical risk function. 
Assume we have a set of training instances 
$S =\{ (x_1, y_1), \cdots, (x_n, y_n)\} \in (\mathcal{X} \times \mathcal{Y})^n$. 
The risk is measured by an user-specified empirical loss function $\Delta(y_i, \hat{y}_i)$, which quantifies the difference between an estimator $\hat{y}_i$ and the correct output $y_i$. 
It is usually difficult to exactly minimize the loss function because it is typically non-convex and discontinuous with $w$ (e.g., Hamming loss). Instead, one adopts surrogate upper bounds to overcome this difficulty.

Assume $\hat{y}_i(w)$ is the marginal MAP prediction on instance $x_i$ as defined in \eqref{e3}. 
We upper bound the empirical loss function $\Delta(y_i, \hat{y}_{i}(w))$ as follows,
\begin{eqnarray}
&&\hspace{-3.5em}\quad \Delta(y_i, \hat{y}_{i}(w)) \nonumber\\
&&\le \Delta(y_i, \hat{y}_{i}(w))
+ \log\sum_{h}\exp[w^{T}\phi(x_i, \hat{y}_{i}(w)), h)] \nonumber\\
&&\qquad - \log\sum_{h}\exp[w^{T}\phi(x_i, {y}_{i}, h)] \quad \quad \nonumber\\
&&\le \max_{y} \Big\{ \Delta(y_i, y) + \log\sum_{h}\exp \ [w^{T}\phi(x_i, y, h)] \Big\} \nonumber\\
&&\qquad - \log\sum_{h}\exp \ [w^{T}\phi(x_i, {y}_{i}, h)], \ \ \nonumber
\end{eqnarray}
where the first inequality holds because $\hat{y}_i(w)$ is the marginal MAP prediction (\ref{e3}), and the second because it jointly maxmizes two terms. 

Minimizing this upper bound over the training set with a $L_2$ regularization, we obtain the following objective function for our marginal structured SVM,

\vspace{-1.5em}{\small
\begin{align}
\frac{1}{2}\|w\|^2 &+ C\sum_{i=1}^{n}\max_{y} \Big\{ \Delta(y_i, y) + \log\sum_{h} \exp[w^{T}\phi(x_i, y, h)] \Big\} \nonumber \!\!\!\!\\
& - C\sum_{i=1}^{n}\log\sum_{h}\exp \Big[w^{T}\phi(x_i, {y}_{i}, h) \Big].
\label{e5}
\end{align} 
}
The constraint form of (\ref{e5}) can be found in the supplement. Note that the first part of the objective requires a loss-augmented \emph{marginal MAP} inference, which marginalizes the hidden variables $h$ and then optimizes over the output variables $y$, while the second part only requires a marginalization over the hidden variables. Both these terms and their gradients are intractable to compute on loopy graphical models, but can be efficiently approximated by mixed-product belief propagation \citep{Liu13} and sum-product belief propagation \citep{Wainwright08}, respectively.
We will discuss training algorithms for optimizing this objective in Section~\ref{sec:learning}. 
\section{A Unified Framework}
\label{sec:Unified Framework}
In this section, we compare our framework with a spectrum of existing methods, and introduce a more general framework that includes all these methods as special cases. 
To start, note that the objective function of the LSSVM \citep{Joachims09} is 
\begin{align} 
\ \frac{1}{2}\|w\|^2  & + C\sum_{i=1}^{n}\max_{y}\max_{h} \Big\{ \Delta(y_i, y) +  w^{T}\phi(x_i, y, h)\Big\} \nonumber \\
&  - C\sum_{i=1}^{n} \max_{h} \Big[ w^{T}\phi(x_i, {y}_{i}, h) \Big].
\label{e6}
\end{align}
Our objective in \eqref{e5} is similar to \eqref{e6}, except replacing the \emph{max} operator of $h$ with the $\log$-$\mathrm{sum}$-$\exp$ function, the so called \emph{soft-max} operator. 
One may introduce a ``temperature" parameter that smooths between \emph{max} and \emph{soft-max}, which motivates a more general objective function that includes MSSVM, LSSVM and other previous methods as special cases, 
\begin{align}
\frac{1}{2}\|w\|^2 &+ C\sum_{i=1}^{n} \epsilon_{y}\log\sum_{y}\exp \ 
 \Big\{ \frac{1}{\epsilon_{y}} \Big[ \Delta(y_i, y) \nonumber\\ 
&+ \ \epsilon_{h}\log\sum_{h}\exp \Big( \frac{w^{T}\phi(x_i, y, h)}{\epsilon_{h}} \Big) \Big] \Big\} \nonumber \\
&- C\sum_{i=1}^{n} \epsilon_{h} \log\sum_{h}\exp \Big(\frac{w^{T}\phi(x_i, y_i, h)}{\epsilon_h}\Big), 
\label{e9}
\end{align}
where $\epsilon_y$ and $\epsilon_h$ are temperature parameters that control how much uncertainty we want account for in $y$ and $h$, respectively.  Similar temperature-based approaches have been used both in structured prediction
\citep{Hazan10,Schwing12} and in other problems, such as semi-supervised learning \citep{Roth12, Dhillon12}.

One can show (Lemma 1 in supplement) that objective~(\ref{e9}) is an upper bound of the empirical loss function 
$\Delta(y_i, \hat{y_i}^{\epsilon_h}(w))$ over the training set, where the prediction $\hat{y_i}^{\epsilon_h}(w)$ is decoded by ``annealed" marginal MAP, 
$$
\hat{y_i}^{\epsilon_h}(w) = \arg\max_{y}  \log\sum_{h}\exp \Big[ \frac{w^{T}\phi(x_i, y, h)}{\epsilon_h}  \Big].
$$

This framework includes a number of existing methods as special cases. 
It reduces to MSSVM in \eqref{e5} if $\epsilon_y \to 0^+$ and $\epsilon_h = 1$, and LSSVM in \eqref{e6} if $\epsilon_y\to 0^+$ and $\epsilon_h \to 0^+$. 
If we set $\epsilon_y = \epsilon_h = 1$, we obtain the loss-augmented likelihood objective in \citet{Volkovs11}, and further reduces to the standard likelihood objective of HCRFs if we assume $\Delta(y_i, y) \equiv 0$.
Our framework also generalizes the $\epsilon$-extension model by \citet{Schwing12}, which corresponds to the restriction that $\epsilon_y = \epsilon_h$.
See Table \ref{models-table} for a summarization of these model comparisons. 
In the sequel, we provide some general insights on selecting among these different models through our empirical evaluations. 
\begin{table}[t] \centering
\caption{Model comparisons within our unified framework.}
\tabcolsep 1.5mm
        \begin{tabular}{ l | l  l   }
        \hline
         Model & $\epsilon_{h}\to 0^{+}$($\max_h$) & $\epsilon_{h}=1$ ($\sum_h$) \\ \hline
            $\epsilon_{y}\to 0^{+}$ ($\max_y$) & LSSVM & MSSVM \\ 
            $\epsilon_{y}=1$ ($\sum_y$)       & N/A & HCRF  \\ 
            $\epsilon_{y}=\epsilon_{h} \in (0, 1)$ & \multicolumn{2}{c}{$\epsilon$-extension model}\\ \hline
        \end{tabular}
\label{models-table}
\end{table}
\section{Training Algorithms}
\label{sec:learning}
In this section, we introduce two optimization algorithms for minimizing the objective function in \eqref{e5}, including a sub-gradient descent (SGD) algorithm and a  concave-convex procedure (CCCP).  An empirical comparison of these two algorithms is given in the experiments of Section~\ref{sec:experiments}. 
\subsection{Sub-gradient Descent (SGD)}
According to Danskin's theorem, the sub-gradient of the MSSVM objective \eqref{e5} is:
\begin{align}
\nabla_{w} M = w  &+  C\sum_{i=1}^{n} \E_{p(h|x_i, \hat{y}_i)}[\phi(x_i, \hat{y}_i, h)] \nonumber\\
&- C\sum_{i=1}^{n} \E_{p(h|x_i, y_i)}[\phi(x_i, y_i, h)], \label{e10} 
\end{align}
where,
\begin{align}
\hat{y}_i = \arg\max_{y\in \mathcal{Y}} & \left\{ \Delta(y_i, y) + \log\sum_{h}\exp [w^T\phi(x_i, y, h)] \right\} 
\label{e11}
\end{align}
is the loss-augmented marginal MAP prediction, which can be approximated via mixed-product belief propagation as described in \citet{Liu13}. The $\E_{p(h|x_i, \hat{y}_i)}$ and $\E_{p(h|x_i, y_i)}$ denote the expectation over the distributions $p(h|x_i, \hat{y}_i)$ and $p(h|x_i, y_i)$, respectively. 
Both expectations can similarly be approximated using the marginal probabilities obtained from belief propagation. See Algorithm~\ref{alg:SGD} for details of the sub-gradient descent (SGD) algorithm for MSSVM. 

Furthermore, one can show (Lemma 2 in supplement) that the (sub-)gradient of the unified framework \eqref{e9} is 
\begin{align}
\nabla_{w} U = w   &+   C\sum_{i=1}^{n} \E_{p^{(\epsilon_y, \epsilon_h)} (y, h| x_i)}[\phi(x_i, y, h)]\nonumber\\ 
 &-  C\sum_{i=1}^{n} \E_{p^{\epsilon_h}(h|x_i, y_i)}[\phi(x_i, y_i, h)].
\label{e12}
\end{align}
where the corresponding temperature controlled distributions are defined as,
\begin{align}
p^{\epsilon_h} (h|x_i, y) &\propto \exp \Big[ \frac{w^{T}\phi(x_i, y, h)}{\epsilon_h} \Big] , \nonumber \\
p^{(\epsilon_y, \epsilon_h)} (y|x_i) &\propto \exp \Big\{ \frac{1}{\epsilon_y} \big[ \Delta(y, y_i) \nonumber\\
 & \quad 
 + \epsilon_h \log\sum_{h} \exp\big( \frac{ w^{T}\phi(x_i, y, h) }{ \epsilon_h} \big)  \big] \Big\}  , \nonumber\\
p^{(\epsilon_y, \epsilon_h)} (y, h|x_i) &= p^{\epsilon_h} (h|x_i, y) \cdot p^{(\epsilon_y, \epsilon_h)} (y|x_i). \nonumber
\end{align}
Exactly as in Table \ref{models-table},  this reduces to the sub-gradient of MSSVM \eqref{e10} if  $\epsilon_y \to 0^+$ and $\epsilon_h = 1$,  the sub-gradient of LSSVM if $\epsilon_y \to 0^+$ and $\epsilon_h \to 0^+$, and the gradient of HCRF if $\epsilon_y = 1$, $\epsilon_h = 1$ and $\Delta(y, y_i) \equiv 0$.
One can simply substitute these (sub-)gradients into Algorithm~\ref{alg:SGD} to obtain the corresponding training algorithms for LSSVM and HCRF.  In those cases, max-product BP and sum-product BP can be used to approximate the inference operations instead.

\begin{algorithm}[tb]
   \caption{Sub-gradient Descent for MSSVM}
   \label{alg:SGD}
\begin{algorithmic}
   \STATE {\bfseries Input:} 
   		number of iterations $T$, learning rate $\eta$
   \STATE {\bfseries Output:} 
     	the learned weight vector $w^*$
   \STATE $w = 0$	  
   \FOR{$t=1$ {\bfseries to} $T$}
   \STATE $\nabla_{w} = 0$
   \FOR{$i=1$ {\bfseries to} $n$}
   \STATE 1. Calculate $\phi_m = \E_{p(h|x_i, \hat{y}_i)}[\phi(x_i, \hat{y}_i, h)]$ by mixed-product BP ($\hat{y_i}$ is defined in (\ref{e11}))
   \STATE 2. Calculate $\phi_s = \E_{p(h|x_i, y_i)}[\phi(x_i, y_i, h)]$ by sum-product BP
   \STATE 3.  \ \ $\nabla_{w} \leftarrow \nabla_{w} + C  (\phi_m - \phi_s)$
   \ENDFOR
   \STATE $w \leftarrow (1-\eta) w - \eta \nabla_{w}$
   \ENDFOR
   \STATE $w^* \leftarrow w$
\end{algorithmic}
\end{algorithm}
\subsection{CCCP Training Algorithm}
The concave-convex procedure (CCCP) \citep{Yuille03} is a general non-convex optimization algorithm with wide application in machine learning. 
It is based on the idea of rewriting the non-convex objective function into a sum of a convex function and a concave function (or equivalently a difference of two convex functions), and transforming the non-convex optimization problem into a sequence of convex sub-problems by linearizing the concave part. 
CCCP provides a straightforward solution for our problem, since the objective functions of all the methods we have discussed -- in \eqref{e5}, \eqref{e6} and \eqref{e9} -- are naturally differences of two convex functions. For example, the MSSVM objective in \eqref{e5} can be naturally written as,
\begin{align}
 f(w) &= f^+(w) - f^{-}(w), \qquad \qquad \qquad  \nonumber
\end{align}
where,
\begin{align}
f^{+}(w) &= \frac{1}{2}\|w\|^2   + C\sum_{i=1}^{n}\max_{y} \Big\{ \Delta(y_i, y)  \nonumber \\
 &\qquad + \log\sum_{h}\exp[w^{T}\phi(x_i, y, h)]\Big\} , \nonumber\\
f^{-}(w) &= C\sum_{i=1}^{n}  \log\sum_{h}\exp[w^{T}\phi(x_i, y_i, h)]. \nonumber
\end{align}
Denoting the parameter vector at iteration $t$ by $w^t$, 
the CCCP algorithm updates to new parameters $w^{t+1}$ by minimizing a convex 
surrogate function where $f^{-}(w)$ is linearized:
\begin{multline*}
w^{t+1}  \gets \arg \min_{w} \{  f^{+}(w)  -   w ^T  \nabla f^{-}(w^t) \} ,  \\
\text{where ~~}\nabla f^{-}(w^t) =   C\sum_i \E_{p(h | x_i, y_i)} [\psi(x_i, y_i, h)]
\end{multline*}
is the gradient of $f^{-}(w)$ at $w_t$ and its expectation 
can be evaluated (approximately) by belief propagation. 
See Algorithm~\ref{alg:CCCP} for more details of CCCP for the MSSVM.

\begin{algorithm}[tb]
   \caption{CCCP Training of MSSVM}
   \label{alg:CCCP}
\begin{algorithmic}
   \STATE {\bfseries Input:} 
   		number of outer iterations $T$, learning rate $\eta$, tolerance $\epsilon$ for inner loops
   \STATE {\bfseries Output:} 
     	the learned weight vector $w^*$  
   
   \STATE $w = 0$
   
   \FOR{$t=1$ {\bfseries to} $T$}
	\STATE $u  = 0$   
   \FOR{$i=1$ {\bfseries to} $n$}
   \STATE 1. Calculate $\phi_s = \E_{p(h|x_i, y_i)}[\phi(x_i, y_i, h)]$ by \\ \hspace{1 em} sum-product BP
   \STATE 2. $u = u + \phi_s$
   \ENDFOR
   
	\REPEAT
	\STATE $\nabla_{w} = 0$	
	\FOR{$i=1$ {\bfseries to} $n$}
		 \STATE 1. Calculate $\phi_m = \E_{p(h|x_i, \hat{y}_i)}[\phi(x_i, \hat{y}_i, h)]$ by \\ \hspace{1 em}  mixed-product BP ($\hat{y_i}$ is defined in (\ref{e11}))
   		\STATE 2. $\nabla_{w} \leftarrow \nabla_{w} + C  \phi_m$
	\ENDFOR
	\STATE $\nabla_{w} = \nabla_{w} - C  u$
	\STATE $w \leftarrow (1-\eta) w - \eta \nabla_{w}$
	\UNTIL{$|| \nabla_w|| \leq \epsilon$ }
   \ENDFOR
   \STATE $w^* \leftarrow w$
\end{algorithmic}
\end{algorithm}

\section{Experiments} \label{sec:experiments}
In this section, we compare our MSSVM with other state-of-art methods on both simulated and real-world datasets.
We demonstrate that the MSSVM significantly outperforms LSSVM, max-margin min-entropy (M3E) model \citep{Miller12} and loss-based learning by modeling latent variable(ModLat) \citep{Kumar12}, especially when the uncertainty over hidden variables is high. Our method also largely outperforms HCRFs in all experiments, especially with a small training sample size.
\subsection{Simulated Data}
\begin{figure}[t]
\vskip 0.2in
\centering
\begin{tabular}{cc}
\!\!{\includegraphics[width=.22\textwidth]{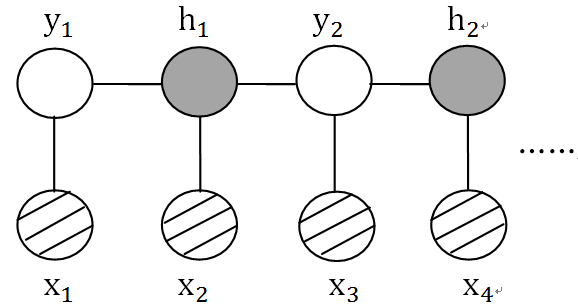}}
& 
{\includegraphics[width=.22\textwidth]{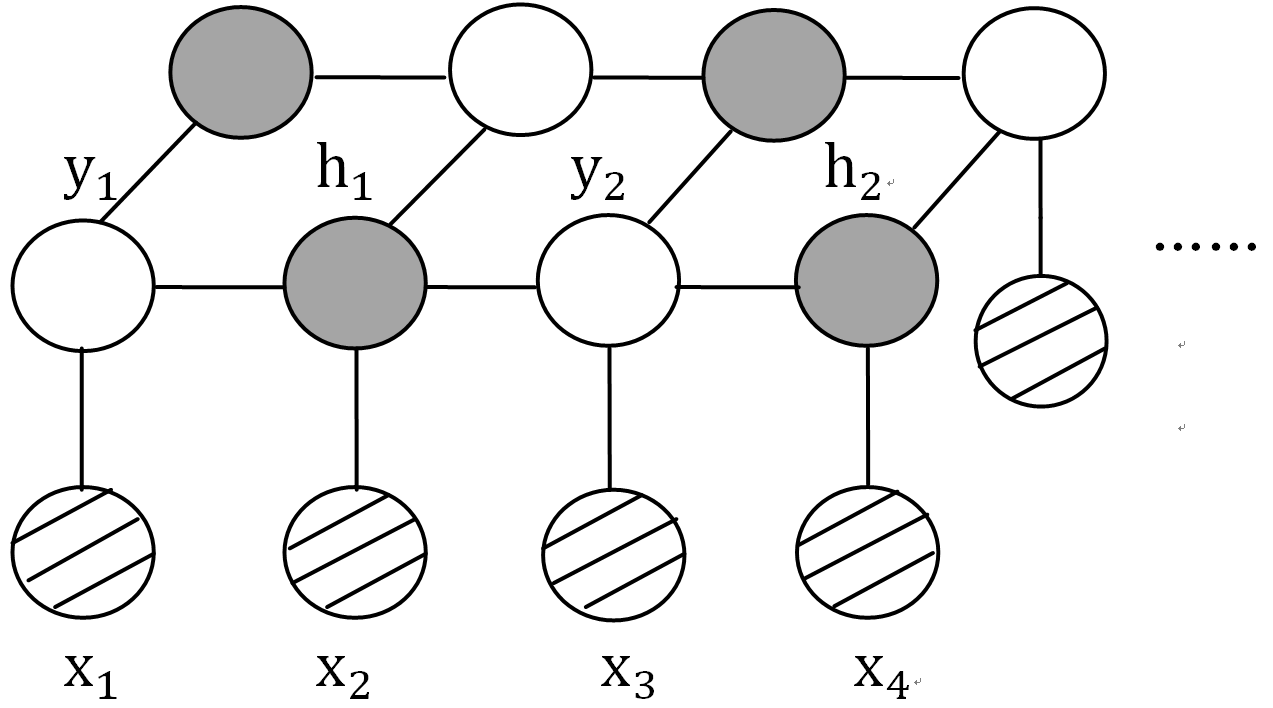}} \\
{\small (a)} &  {\small (b)}
\end{tabular}
\caption{(a) The hidden chain and (b) 2D grid model used in our simulation experiments. The shaded nodes denote hidden variables $h$,  while the unshaded nodes are the output variables $y$ and nodes with hatching are the inputs $x$.}
\label{fig:hand_draw}
\end{figure}

We simulate both training and testing data from a pairwise Markov random field (MRF)  over graph $G = (V, E)$ with discrete random variables taking values in $\{0, 1, 2, 3\}^n$, given by,
\begin{align}
&p(x, y, h|w)\propto \nonumber\\ 
&\exp \Big[ \sum_{x_i \in V} w_{x_i}^T \phi({x_i}) + \sum_{y_j \in V} w_{y_j}^T \phi({y_j}) + \sum_{h_k \in V} w_{h_k}^T \phi({h_k}) \nonumber \\
&+ \sum_{(x_i, y_j) \in E} \!\!\!\! w_{(x_i,y_j)}^T \phi(x_i, y_j) + \sum_{(x_i, h_k) \in E} \!\!\!\! w_{(x_i,h_k)}^T \phi(x_i, h_k) \nonumber \\
\ &+ \sum_{(y_j, h_k) \in E} \!\!\!\! w_{(y_j,h_k)}^T \phi(y_j, h_k) \Big] ,  \nonumber
\label{sim_model}
\end{align}
where the graph structure G is either a ``hidden chain'' (40 nodes) or a 2D grid (size $6\times 6 \times 2 = 72$ nodes), as illustrated in Figure~\ref{fig:hand_draw}. The log-linear weights $w$ are randomly generated from normal distributions.
\newcommand{\N}{{N}}
The singleton parameters $w_{x_i} $, $w_{y_j}$ and $w_{h_k}$ are drawn from $\N(0,\ \sigma_x^2 \cdot I)$, $\N(0,\ \sigma_y^2 \cdot I)$ and  $\N(0, \sigma_h^2 \cdot I)$, respectively, corresponding to indicator vectors $\phi(x_i)$, $\phi(y_j)$ and $\phi(h_k)$. 
The pairwise parameters $w_{(y_j, h_k) = (s, t)}$, $w_{(x_i, y_j) = (r, s)} $ and $w_{(x_i, h_k) = (r, t)} $ are drawn from $\N(0, \sigma_{yh}^2)$,  $\N(0, \sigma_{xy}^2)$ and $\N(0, \sigma_{xh}^2)$, respectively, corresponding to indicators $\phi(y_j=s,\ h_k=t)$, $\phi(x_i=r,\ y_j=s)$ and $\phi(x_i=r,\ h_k=t)$. 
Note that the variance parameters $\sigma_h$ and $\sigma_{yh}$ control the degree of uncertainty in the hidden variables and their importance for estimating the output variables $y$:
the uncertainty of $h$ is high for small values of $\sigma_h$, and the correlation between $h$ and $y$ is high when $\sigma_{yh}$ is large. 

We sample 20 training instances and 100 test instances from both the hidden chain MRF and 2D grid MRF. We set $\sigma_{x} = \sigma_{y} = \sigma_{h} = 0.1$, $\sigma_{yh} = \sigma_{yx} = \sigma_{hx} = 2$. Then, we train our MSSVM, LSSVM and HCRF models using both SGD and CCCP. Hamming loss is used in both training and evaluation. In our experiments, we always set the regularization weight $C = 1$.   See Table~\ref{tab:SGD_vs_CCCP} for the results across different algorithms.
We can see that our MSSVM always achieves the highest accuracy when using either training algorithm.
It is worth noting that LSSVM obtains a significantly better result using CCCP than SGD;
this is mainly due to SGD's difficulty converging on the piecewise linear objective of LSSVM.
\begin{table}[t] 
\tabcolsep 1.5mm
\caption{Average accuracy (\%) of MSSVM, LSSVM, HCRFs using SGD and CCCP when the data are simulated from $40$-node hidden chain and $6\times6$ 2D-grid graph as shown in Figure~\ref{fig:hand_draw}. 
The results are averaged over 20 random trials. }
\begin{center}
    \begin{tabular}{c|ccc}
    \hline
    Hidden Chain   & MSSVM  & LSSVM   & HCRFs  \\\hline
    SGD & ${\bf 69.20}$  & $66.87$  & $68.75$ \\
    CCCP & ${\bf 69.63}$ &$ 67.91$ & $ 69.03 $ \\ \hline
    \end{tabular}
\end{center}
\begin{center}
    \begin{tabular}{c|ccc}
    \hline
    2D-grid graph & MSSVM  & LSSVM   & HCRFs  \\ \hline
    SGD & ${\bf 74.12}$  & $71.96$  & $73.51$ \\
    CCCP & ${\bf 74.08}$ &$ 73.38$ & $ 73.62 $ \\ \hline
    \end{tabular}
\end{center}
\label{tab:SGD_vs_CCCP}
\end{table}
\begin{figure}[t] \centering
\label{fig:convergence}
\centerline{\includegraphics[width=6.2cm]{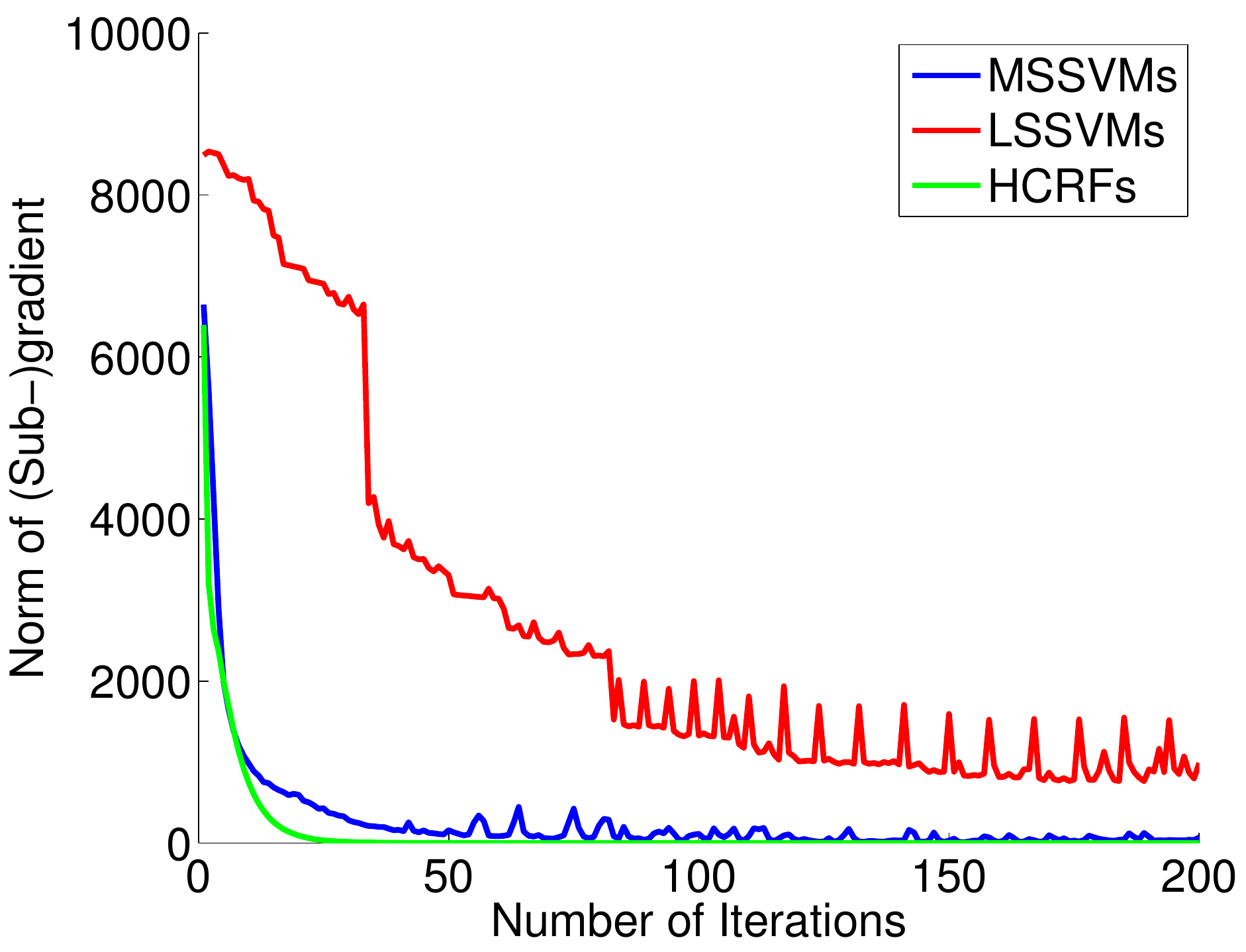}}
\caption{Convergence behavior of (sub-)gradient descent on MSSVMs, LSSVMs and HCRFs. }
\end{figure}
\paragraph{Empirical Convergence of SGD and CCCP.}
Using sub-gradient descent with learning rate $\eta_M = 0.02$, we found that for our MSSVM, training error converged quickly (within 50 iterations).
However, sub-gradient descent on the LSSVM would only converge using a much smaller learning rate ($\eta_L = 0.001$), and converged more slowly (usually after $250$ iterations).
This effect is mainly because the LSSVM hard-max makes the objective function nonsmooth, causing sub-gradient descent to be slow to converge.
On the other hand, gradient descent on HCRFs converges more easily and quickly than either MSSVM or LSSVM, because its objective function is smoother.
Figure~2 shows the oscillation during the iteration of  (sub-)gradient descent for each model, and empirically illustrates the convergence process.

We also observe CCCP converging faster than SGD (using smaller number of inference steps), especially for LSSVM, since CCCP transforms the complex piecewise linear objective into a sequence of easier convex sub-problems.
In our empirical study, CCCP always converged well even using approximate inference and non-convex objectives.\footnote{However, it is challenging to provide rigorous convergence guarantees for the non-convex \& intractable setting, and not really the focus of this paper.}
To provide a fair comparison, all methods are trained using the CCCP algorithm in the sequel.
%
%
\paragraph{Training Sample Size.}
We compared the influence of sample size for each method by ranging the training size from $2^2$ to $2^{10}$ (with a testing size of $500$).  The data are all simulated from a MRF on the 20-node hidden chain shown in Figure~\ref{fig:hand_draw}(a). We set $\sigma_{x} = \sigma_{y} = \sigma_{h} =  0.1$ and $\sigma_{yh} = \sigma_{yx} = \sigma_{hx} = 2$ as before.
\begin{figure}[t] \centering
\centerline{\includegraphics[width=6.5cm]{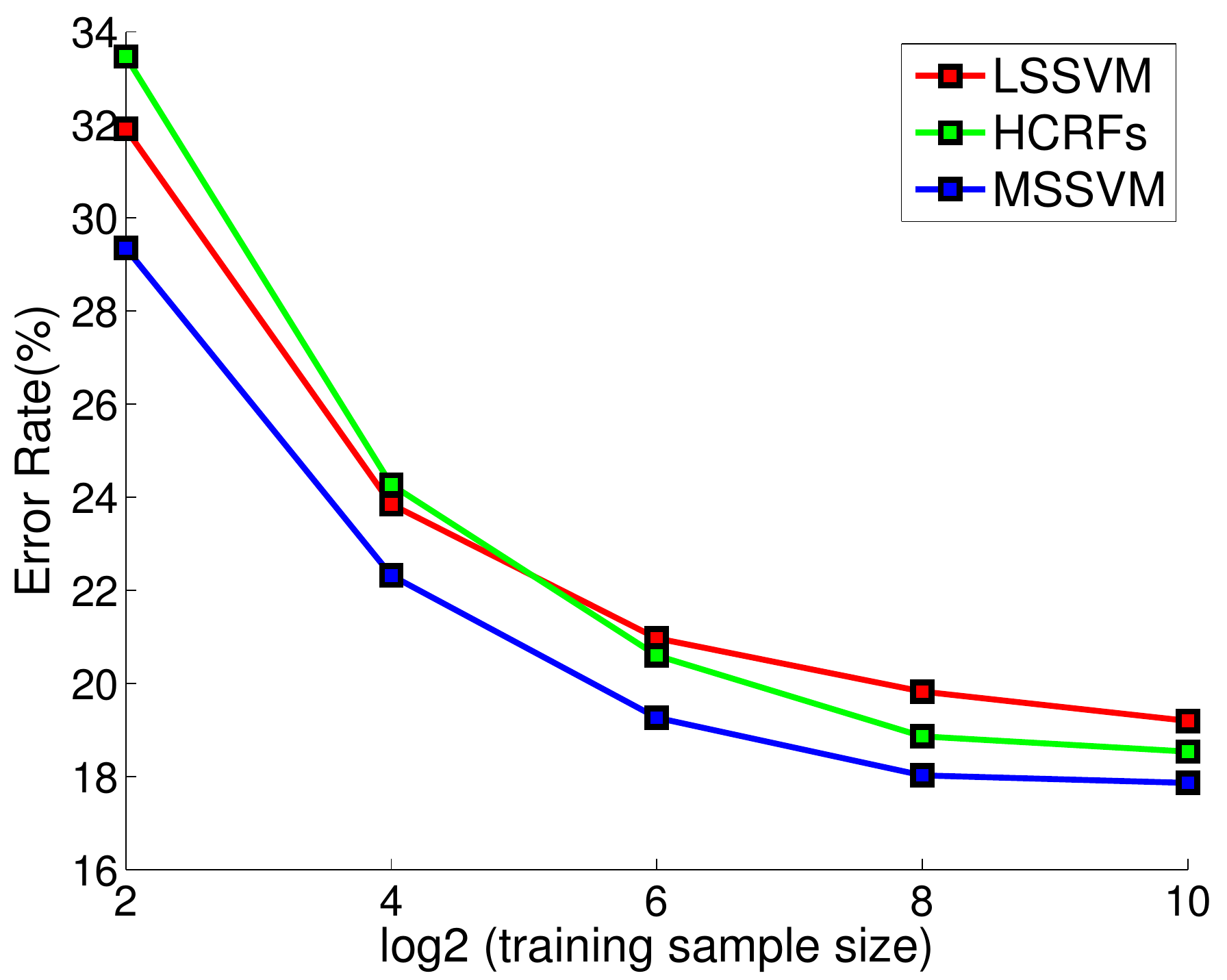}}
\caption{ The error rate of MSSVM, LSSVM and HCRFs as the training sample size increases. Results are averaged over 5 random trials.}
\label{fig:training_size}
\end{figure}

Results are shown in Figure~3. We found that our MSSVM always considerably outperforms LSSVM, and largely outperforms HCRFs when the training sample sizes are small. HCRFs perform worse than LSSVM for few training data, but outperform LSSVM as the training sample increases. 

Our experiment shows that MSSVM consistently outperforms HCRFs even with reasonably large training sets on a relatively simple toy model. 
Although the maximum likelihood estimator (as used in HCRFs) is generally considered asymptotically optimal if the model assumptions are correct, this assumes a sufficiently large training size, which may be difficult to acheive in practice.  
Given enough data (and the correct model), the HCRF should thus eventually improve, but this seems unrealistic
in practice
since most applications are likely to exhibit high dimensional parameters and relatively few training instances. 
Additional analysis of the test likelihood and prediction accuracy can be found in the supplement.
%
\paragraph{Uncertainty of Hidden Variables.} 
We investigate the influence of uncertainty in the hidden variables for each method by adjusting the noise level $\sigma_h$, which controls the uncertainty of the hidden variables. We draw 20 training samples and 100 test samples from a MRF on a 40-node hidden chain shown in Figure~\ref{fig:hand_draw}(a), with fixed  $\sigma_{x} = \sigma_{y} = 0.1$ and $\sigma_{yh} = \sigma_{yx} = \sigma_{hx} = 2$.  For comparison, we also evaluate the performance of M3E \citep{Miller12} and ModLat \citep{Kumar12}. 
In accordance with our default setting $C = 1$, we use the default hyper-parameters in their package. We encourage people to carefully tune these hyper-parameters by cross-validation in future study.

Results are shown in Table~\ref{tab:sigma_h}.  We find that our MSSVM is competitive with LSSVM and M3E when the uncertainty in the hidden variables is low, and becomes significantly better than them as the uncertainty increases. Because LSSVM uses the joint MAP, it does not take into account this uncertainty. On the other hand, M3E explicitly tries to minimize this uncertainty, which can also mislead the prediction.
Our MSSVM consistently outperforms HCRFs for moderate training sample sizes. 
Due to the limitations in current implementations of M3E and ModLat, we only provide their results on chain models.  

\begin{table}[t]
\caption{The accuracy (\%) of MSSVM, LSSVM, HCRFs, M3E and ModLat under different $\sigma_h$, which governs the level of uncertainty in the hidden variables.  Small values of $\sigma_h$ correspond to high uncertainty in hidden variables. Results are averaged over 20 random trials.}
\tabcolsep 1.5mm
    \begin{center}
        \begin{tabular}{ l | l  l  l  l  l}
        \hline
         $\sigma_h$ &  MSSVM   & LSSVM &   HCRFs   &  M3E\quad\ &  ModLat \\ \hline
            10 &    ${79.30}$   & ${\bf 79.46}$ & $78.68$ & $79.04$ &  $77.16$ \\ 
            1 &     ${70.00}$   &${\bf 70.07}$  & $69.88$  & $68.53$ &  $67.91$ \\ 
            0.5 &   ${\bf 67.24}$   &$65.98$   & $66.66$  & $66.05$ &  $65.15$ \\ 
            0.1 &   ${\bf 69.63}$   &$67.91$   & $69.03$  & $65.19$ &  $67.96$ \\ 
            0.01 &  ${\bf 73.88}$   &$71.38$   & $72.58$ & $67.21$ &  $71.52$ \\ 
            1e-3 &  ${\bf 72.08}$   &$69.24$   & $70.88$ & $65.48$ & $66.54$ \\ \hline
            Avg. & ${\bf 72.02}$    &$70.67$    & $71.28$ & $68.58$ & $69.37$ \\ \hline
        \end{tabular}
    \end{center}
\label{tab:sigma_h}    
\end{table}
\begin{figure*}[th] \centering
\begin{tabular}{ccc}
\!\!\includegraphics[width=5.2cm, height=3.6cm, clip]{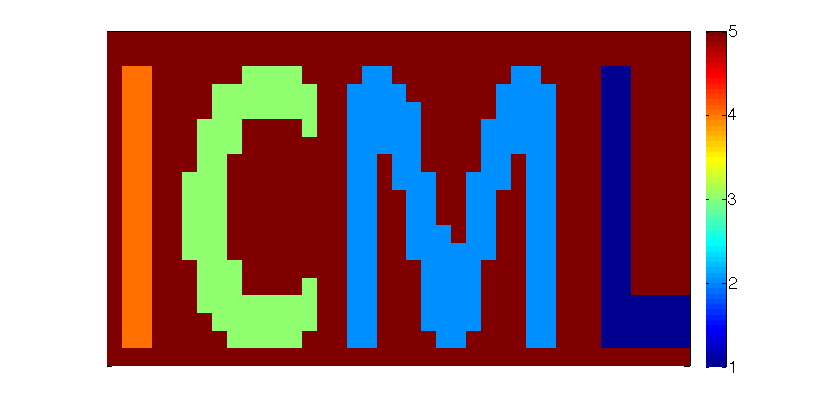} \!\!&\!\! 
\includegraphics[width=5.2cm, height=3.6cm, clip]{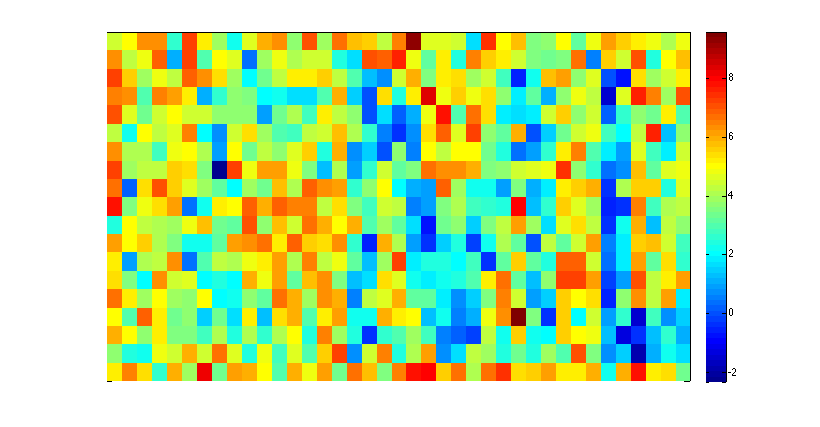} \!\!&\!\! 
\includegraphics[width=5.7cm, height=4.6cm, clip]{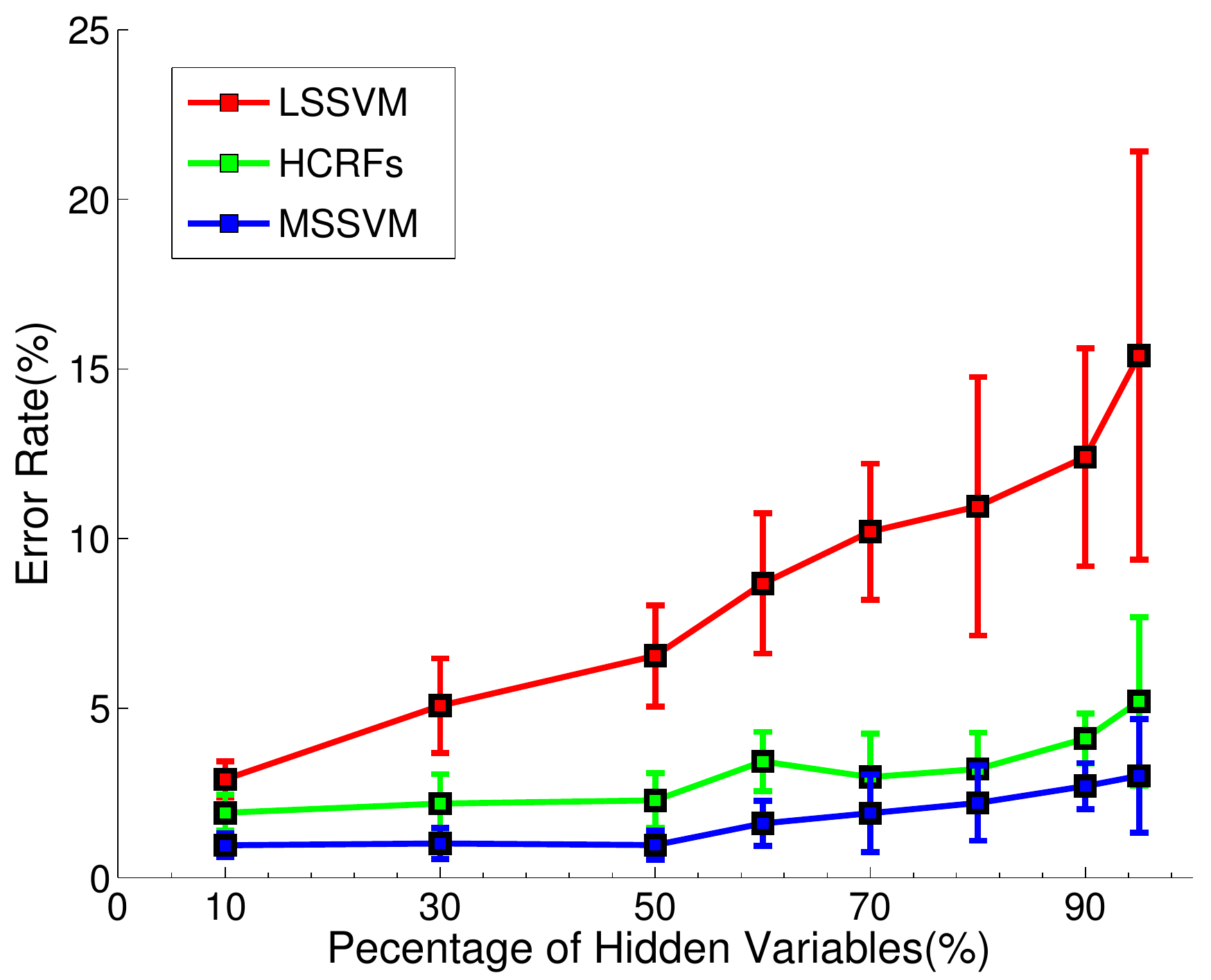} \\ 
{\small(a) }& {\small(b)}  & {\small(c)}
\end{tabular}
\caption{ (a) The ground truth image. (b) An example of an observed noisy image. (c) The performance of each algorithm as the percentage of missing labels varies from $10\%$ to $95\%$. Results are averaged over 5 random trials, each using 10 training instances and 10 test instances. }
\label{fig:f4} %
\end{figure*}
\subsection{Image Segmentation}
In this section, we evaluate our MSSVM method on the task of segmenting weakly labeled images. 
Our settings are motivated by the experiments in \citep{Schwing12}. 
We assume a ground truth image of 20$\times$40 pixels as shown in Figure~\ref{fig:f4} (a), where each pixel $i$ has a label $y_i$ taking values in $\{1, \cdots, 5\}$.  The observed image $x$ is obtained by  adding Gaussian noise, $\N(0, 5)$, on the ground truth image as Figure~\ref{fig:f4} (b). 

We use the 2D-grid model as in Figure~\ref{fig:hand_draw} (b), with local features $\phi(y_i, x_i) = e_{y_i} \otimes x_i$ and pairwise features $\phi(y_i, y_j) = e_{y_i} \otimes e_{y_j} \in \R^{5\times 5}$ as defined in \citet{Nowozin11}, where $e_{y_i}$ is the unit normal vector with entry one on dimension $y_i$ and $\otimes$ is the outer product. The set of missing labels  (hidden variables) are determined at random, in proportions ranging from $10\%$ to $95\%$.  The performance of MSSVM, LSSVM, and HCRFs are evaluated using the CCCP algorithm.

Figure~\ref{fig:f4} (c) lists the performance of each method as the percentage of missing labels is increased. We can see that the performance of LSSVM degrades significantly as the number of hidden variables grows. Most notably, MSSVM is consistently the best method across all settings. 
This can be explained by the fact that the MSSVM combines both the max-margin property
and the improved robustness given by properly accounting for uncertainty in the hidden labels.

\subsection{Object Categorization}
Finally, we evaluate our MSSVM method on the task of object categorization using partially labeled images. We use the Microsoft Research Cambridge data set \citep{Winn05}, consisting of 240 images with 213$\times$320 pixels and their partial pixel-level labelings. 
The missing labels may correspond to ambiguous regions, undefined categories or object boundaries, etc.

Modeled on the approach outlined in \citet{Verbeek06},
we use $20\times20$ pixel patches with centers at 10 pixel intervals and treat each patch as a node in our model. This results in a $20\times31$ grid model as in Figure~\ref{fig:hand_draw} (b). 
The local features of each patch are encoded using texture and color descriptors. For texture, we compute the 128-dimensional SIFT descriptor of the patch and vector quantize it into a 500-word codebook, learned by k-means clustering of all patches in the entire dataset. For color, we take 48-dimensional RGB color histogram for each patch. In our experiment, we select the 5 most frequent categories in the dataset and use 2-fold cross validation for testing.

Table \ref{tab:MSRC} shows the accuracies of each method across the various categories. 
Again, we find that MSSVM consistently outperforms other methods across all categories, which can be explained by both the superiority of SSVM-based methods for moderate sample size and the improved robustness by maintaining the uncertainty over the missing labels in the learning procedure.
\begin{table}[t]
\tabcolsep 1.5mm
\caption{Average patch level accuracy (\%) of MSSVM, LSSVM, HCRFs for MSRC data by 2-fold cross validation.}
\begin{center}
    \begin{tabular}{c|ccc}
    \hline
    MSRC Data  & MSSVM  & LSSVM   & HCRFs  \\ \hline
    Building & ${\bf 72.4}$  & $70.7$  & $71.7$ \\
    Grass & ${\bf 89.7}$ & $88.9$ & $ 88.3 $ \\ 
    Sky  & ${\bf 88.3}$ & $85.6$ & $ 88.2 $\\ 
    Tree  & ${\bf 71.9}$ & $71.0$ & $ 70.1 $\\  
    Car  & ${\bf 70.8}$ & $69.4$ & $ 70.2 $\\ \hline
    \end{tabular}
\end{center}
\label{tab:MSRC}
\vskip -0.15in
\end{table}
\section{Conclusion}
\label{sec:conclusion}
We proposed a novel structured SVM method for structured prediction with hidden variables. We demonstrate that our MSSVM consistently outperforms state-of-the-art methods in both simulated and real-world datasets, especially when the uncertainty of hidden variables is large. 
Compared to the popular LSSVM, 
the objective function of
our MSSVM is easier to optimize due to the smoothness of its objective function. 
We also provide a unified framework which includes our method as well as a spectrum of previous methods as special cases.

\paragraph{Acknowlegements.}
This work was sponsored in part by NSF grants IIS-1065618 and IIS-1254071, 
and in part by by the United States Air Force under Contract No. FA8750-14-C-0011 under the DARPA PPAML program.

\bibliography{paper_MSSVM}

\begin{thebibliography}{32}
\providecommand{\natexlab}[1]{#1}
\providecommand{\url}[1]{\texttt{#1}}
\expandafter\ifx\csname urlstyle\endcsname\relax
  \providecommand{\doi}[1]{doi: #1}\else
  \providecommand{\doi}{doi: \begingroup \urlstyle{rm}\Url}\fi

\bibitem[Dhillon et~al.(2012)Dhillon, Keerthi, Bellare, Chapelle, and
  and]{Dhillon12}
Dhillon, P., Keerthi, S.~Sathiya, Bellare, K., Chapelle, O., and and,
  S.~Sundararajan.
\newblock Deterministic annealing for semi-supervised structured output
  learning.
\newblock In \emph{Proceedings of AISTAT}, pp.\  299--307, 2012.

\bibitem[Fergus et~al.(2006)Fergus, Singh, Hertzmann, Roweis, and
  Freeman]{Fergus06}
Fergus, R., Singh, B., Hertzmann, A., Roweis, S.~T., and Freeman, W.~T.
\newblock Removing camera shake from a single photograph.
\newblock In \emph{Proceeding of ACM SIGGRAPH}, pp.\  787--794, 2006.

\bibitem[Getoor \& Taskar(2007)Getoor and Taskar]{Taskar07}
Getoor, L. and Taskar, B.
\newblock \emph{Introduction to statistical relational learning}.
\newblock The MIT press, 2007.

\bibitem[Hazan \& Urtasun(2010)Hazan and Urtasun]{Hazan10}
Hazan, T. and Urtasun, R.
\newblock A primal-dual message-passing algorithm for approximated large scale
  structured prediction.
\newblock In \emph{Proceedings of NIPS}, 2010.

\bibitem[Koller \& Friedman(2009)Koller and Friedman]{Koller09}
Koller, D. and Friedman, N.
\newblock \emph{Probabilistic graphical models: principles and techniques}.
\newblock The MIT press, 2009.

\bibitem[Kumar et~al.(2012)Kumar, Packer, and Koller]{Kumar12}
Kumar, P., Packer, B., and Koller, D.
\newblock Modeling latent variable uncertainty for loss-based learning.
\newblock In \emph{Proceedings of ICML}, pp.\  465--472, 2012.

\bibitem[Lafferty et~al.(2001)Lafferty, McCallum, and Pereira]{Lafferty01}
Lafferty, J., McCallum, A., and Pereira, F.
\newblock Conditional random fields: Probabilistic models for segmenting and
  labeling sequence data.
\newblock In \emph{Proceedings of ICML}, pp.\  282--289, 2001.

\bibitem[Levin et~al.(2011)Levin, Weiss, Durand, and Freeman]{Levin11}
Levin, A., Weiss, Y., Durand, F., and Freeman, W.T.
\newblock Efficient marginal likelihood optimization in blind deconvolution.
\newblock In \emph{Proceedings of CVPR}, pp.\  2657--2664, 2011.

\bibitem[Li et~al.(2007)Li, Lin, Wang, and Liu]{Li07}
Li, M.~H., Lin, L., Wang, X.L., and Liu, T.
\newblock Protein--protein interaction site prediction based on conditional
  random field.
\newblock \emph{Bioinformatics}, 23, 2007.

\bibitem[Liu \& Ihler(2013)Liu and Ihler]{Liu13}
Liu, Q. and Ihler, A.
\newblock Variational algorithms for marginal map.
\newblock \emph{JMLR}, 14:\penalty0 3165--3200, 2013.

\bibitem[Miller et~al.(2012)Miller, Kumar, Packer, Goodman, and
  Koller]{Miller12}
Miller, K., Kumar, P., Packer, B., Goodman, D., and Koller, D.
\newblock Max-margin min-entropy models.
\newblock In \emph{Proceedings of AISTATS}, pp.\  779--787, 2012.

\bibitem[Naradowsky et~al.(2012)Naradowsky, Riedel, and Smith]{Naradowsky12}
Naradowsky, J., Riedel, S., and Smith, D.
\newblock Improving {NLP} through marginalization of hidden syntactic
  structure.
\newblock In \emph{Proceeding of EMNLP}, pp.\  810--820, 2012.

\bibitem[Nowozin \& Lampert(2011)Nowozin and Lampert]{Nowozin11}
Nowozin, S. and Lampert, C.
\newblock Structured prediction and learning in computer vision.
\newblock \emph{Foundations and Trends in Computer Graphics and Vision}, 6,
  2011.

\bibitem[Quattoni et~al.(2004)Quattoni, Collins, and Darrell]{Quattoni04}
Quattoni, A., Collins, M., and Darrell, T.
\newblock Conditional random fields for object recognition.
\newblock In \emph{Proceedings of NIPS}, pp.\  1097--1104, 2004.

\bibitem[Quattoni et~al.(2007)Quattoni, Wang, Morency, Collins, and
  Darrell]{Quattoni07}
Quattoni, A., Wang, S., Morency, L., Collins, M., and Darrell, T.
\newblock Hidden conditional random fields.
\newblock \emph{IEEE Transactions on PAMI}, 29:\penalty0 1848--1852, 2007.

\bibitem[Ratliff et~al.(2007)Ratliff, Bagnell, and Zinkevich]{Ratliff07}
Ratliff, N., Bagnell, J.~A., and Zinkevich, M.
\newblock ({O}nline) {S}ubgradient methods for structured prediction.
\newblock In \emph{Proceedings of AISTATS}, pp.\  380--387, 2007.

\bibitem[Samdani et~al.(2012)Samdani, Chang, and Roth]{Roth12}
Samdani, R., Chang, M.W., and Roth, D.
\newblock Unified expectation maximization.
\newblock In \emph{Proceedings of NAACL}, pp.\  688--698, 2012.

\bibitem[Sato \& Sakakibara(2005)Sato and Sakakibara]{Sato05}
Sato, K. and Sakakibara, Y.
\newblock {RNA} secondary structural alignment with conditional random fields.
\newblock \emph{Bioinformatics}, 21, 2005.

\bibitem[Schwing et~al.(2012)Schwing, Hazan, Pollefeys, and Urtasun]{Schwing12}
Schwing, A., Hazan, T., Pollefeys, M., and Urtasun, R.
\newblock Efficient structured prediction with latent variables for general
  graphical models.
\newblock In \emph{Proceedings of ICML}, pp.\  959--966, 2012.

\bibitem[Taskar et~al.(2003)Taskar, Guestrin, and Koller]{Taskar03}
Taskar, B., Guestrin, C., and Koller, D.
\newblock Max-margin {M}arkov networks.
\newblock In \emph{Proceedings of NIPS}, 2003.

\bibitem[Tsochantaridis et~al.(2005)Tsochantaridis, Joachims, Hofmann, and
  Altun]{Tsochantaridis05}
Tsochantaridis, I., Joachims, T., Hofmann, T., and Altun, Y.
\newblock Large margin methods for structured and interdependent output
  variables.
\newblock \emph{JMLR}, 6:\penalty0 1453--1484, 2005.

\bibitem[Verbeek \& Triggs(2007)Verbeek and Triggs]{Verbeek06}
Verbeek, J. and Triggs, B.
\newblock Scene segmentation with {CRF}s learned from partially labeled images.
\newblock In \emph{Proceedings of NIPS}, pp.\  1553--1560, 2007.

\bibitem[Volkovs et~al.(2011)Volkovs, Larochelle, and Zemel]{Volkovs11}
Volkovs, M., Larochelle, H., and Zemel, R.~S.
\newblock Loss-sensitive training of probabilistic conditional random fields.
\newblock \emph{Technical report}, 2011.

\bibitem[Wainwright \& Jordan(2008)Wainwright and Jordan]{Wainwright08}
Wainwright, M. and Jordan, M.I.
\newblock Graphical models, exponential families, and variational inference.
\newblock \emph{Foundations and Trends in Machine Learning}, 1--2:\penalty0
  1--305, 2008.

\bibitem[Wang et~al.(2006)Wang, Quattoni, Morency, and Demirdjian]{Wang06}
Wang, S.B, Quattoni, A., Morency, L., and Demirdjian, D.
\newblock Hidden conditional random fields for gesture recognition.
\newblock In \emph{Proceedings of CVPR}, pp.\  1521--1527, 2006.

\bibitem[Wang \& Mori(2009)Wang and Mori]{Wang09}
Wang, Y. and Mori, G.
\newblock Max-margin hidden conditional random fields for human action
  recognition.
\newblock In \emph{Proceedings of CVPR}, pp.\  872--879, 2009.

\bibitem[Winn et~al.(2005)Winn, Criminisi, and Minka]{Winn05}
Winn, J., Criminisi, A., and Minka, T.
\newblock Object categorization by learned universal visual dictionary.
\newblock In \emph{Proceedings of ICCV}, pp.\  1800--1807, 2005.

\bibitem[Xu et~al.(2013)Xu, Rockmore, and Kleinbaum]{Xu13}
Xu, Y., Rockmore, D., and Kleinbaum, A.
\newblock Hyperlink prediction in hypernetworks using latent social features.
\newblock In \emph{Discovery Science}, pp.\  324--339, 2013.

\bibitem[Yessenalina et~al.(2010)Yessenalina, Yue, and Cardie]{Yessen10}
Yessenalina, A., Yue, Y., and Cardie, C.
\newblock Multi-level structured models for document-level sentiment
  classification.
\newblock In \emph{Proceedings of EMNLP}, pp.\  1046--1056, 2010.

\bibitem[Yu \& Joachims(2009)Yu and Joachims]{Joachims09}
Yu, C. and Joachims, T.
\newblock Learning structural {SVM}s with latent variables.
\newblock In \emph{Proceedings of ICML}, pp.\  1169--1176, 2009.

\bibitem[Yuille \& Rangarajan(2003)Yuille and Rangarajan]{Yuille03}
Yuille, A.~L. and Rangarajan, A.
\newblock The concave-convex procedure.
\newblock \emph{Neural Computation}, 15:\penalty0 915--936, 2003.

\bibitem[Zhu et~al.(2010)Zhu, Chen, Yuille, and Freeman]{LongZhu10}
Zhu, L., Chen, Y., Yuille, A., and Freeman, W.
\newblock Latent hierarchical structural learning for object detection.
\newblock In \emph{Proceedings of CVPR}, pp.\  1062--1069, 2010.

\end{thebibliography}
\bibliographystyle{icml2014}
\clearpage


\onecolumn
\icmltitle{Supplement: Marginal Structured SVM with Hidden Variables}

\icmlauthor{Wei Ping}{wping@ics.uci.edu}
\icmlauthor{Qiang Liu}{qliu1@ics.uci.edu}
\icmlauthor{Alexander Ihler}{ihler@ics.uci.edu}
\icmladdress{Department of Computer Science, UC Irvine}

\icmlkeywords{boring formatting information, machine learning, ICML}

\vskip 0.3in

\section*{Constraint Form of Marginal Structured SVM}
Here we give the constraint form of Eq. (3) in the main paper,
\begin{align}
& \min_{w, \{\xi_i \ge 0 \}} \quad \frac{1}{2}\|w\|^2 + C\sum_{i=1}^{n} \xi_i , \\
& \ \text{s.t.} \ \forall i \in \{1,..., n\}, \ \forall y \in \mathcal{Y}, \
\log\sum_{h}\exp \Big[w^{T}\phi(x_i, {y}_{i}, h) \Big] - \log\sum_{h} \exp[w^{T}\phi(x_i, y, h)]
\ge \Delta(y_i, y) - \xi_i ,  \nonumber
\end{align}
where $\{ \xi_i \}$ are the slack variables. One can show that the optimal solution $\{ \xi_i^* \}, w^*$ satisfies,
\begin{align}
\xi_i^* = \max_{y} \Big\{ \Delta(y_i, y) + \log\sum_{h} \exp[w^{*T}\phi(x_i, y, h)] \Big\}  - \log\sum_{h}\exp [w^{*T}\phi(x_i, {y}_{i}, h) ], \nonumber
\end{align}
which gives the same objective value as the the unconstrained form. One can also derive a cutting plane-based training algorithm for this constraint formulation.
\section*{Details of Proofs}
In this section, we give proofs for two lemmas referenced but omitted from the main paper.

\newtheorem*{lemma1}{Lemma 1}
\begin{lemma1}
The objective of the unified framework~(Eq. (5) in main paper) is an upper bound of the empirical loss function 
$\Delta(y_i, \hat{y_i}^{\epsilon_h}(w))$ over the training set, where the prediction $\hat{y_i}^{\epsilon_h}(w)$ is decoded by ``annealed" marginal MAP, 
$$
\hat{y_i}^{\epsilon_h}(w) = \arg\max_{y}  \log\sum_{h}\exp \Big[ \frac{w^{T}\phi(x_i, y, h)}{\epsilon_h}  \Big].
$$
\begin{proof}
\begin{align}
\Delta( y_i, \hat{y_i}^{\epsilon_h}(w) )
\le & \ \Delta(y_i, \hat{y_i}^{\epsilon_h}(w) )
+ \epsilon_h \log\sum_{h}\exp\big[ \frac{ w^{T}\phi( x_i, \hat{y_i}^{\epsilon_h}(w), h) }{\epsilon_h} \big]
- \epsilon_h \log\sum_{h}\exp \big[ \frac{w^{T}\phi(x_i, {y}_{i}, h)}{\epsilon_h} \big] \quad \quad \nonumber\\
\le &\ \epsilon_y \log\sum_{y}\exp \Big\{ \frac{1}{\epsilon_y} \Big[ \Delta(y_i, y) 
 + \epsilon_h \log\sum_{h} \exp  \Big( \frac{w^{T}\phi(x_i, y, h)}{\epsilon_h} \Big) \Big] \Big\} 
 - \epsilon_h \log\sum_{h}\exp \big[ \frac{w^{T}\phi(x_i, {y}_{i}, h)}{\epsilon_h} \big] ,  \nonumber
\end{align}
where the first inequality holds by the definition of $\hat{y_i}^{\epsilon_h}(w)$, and the second holds for $\forall \epsilon_y > 0$, because the summation over $y$  contains $\hat{y_i}^{\epsilon_h}(w)$.
\end{proof}
\end{lemma1}

For convenience, we denote this upper bound as
\begin{align}
U_{i}(w; \epsilon_y, \epsilon_h) &= U_{i}^{+}(w; \epsilon_y, \epsilon_h) - U_{i}^{-}(w; \epsilon_h) \label{Eqn1} \\
\intertext{where} 
U_{i}^{+}(w; \epsilon_y, \epsilon_h) &= \epsilon_y \log\sum_{y}\exp \Big\{ \frac{1}{\epsilon_y} \Big[ \Delta(y_i, y)
 + \epsilon_h \log\sum_{h} \exp  \Big( \frac{w^{T}\phi(x_i, y, h)}{\epsilon_h} \Big) \Big] \Big\} \nonumber\\
U_{i}^{-}(w; \epsilon_h) &= \epsilon_h \log\sum_{h}\exp \big[ \frac{w^{T}\phi(x_i, {y}_{i}, h)}{\epsilon_h} \big].  \nonumber
\end{align}

\newtheorem*{lemma2}{Lemma 2}
\begin{lemma2}
The (sub-)gradient of $U_i(w; \epsilon_y, \epsilon_h)$ in (\ref{Eqn1}) is,
\begin{align}
\nabla_{w} U_i(w; \epsilon_y, \epsilon_h) &=  \E_{p^{(\epsilon_y, \epsilon_h)} (y, h| x_i)}[\phi(x_i, y, h)] 
 -  \E_{p^{\epsilon_h}(h|x_i, y_i)}[\phi(x_i, y_i, h)], \nonumber
\end{align}
where the corresponding temperature controlled distribution is defined as,
\begin{align}
p^{\epsilon_h} (h|x_i, y) &\propto \exp \Big[ \frac{w^{T}\phi(x_i, y, h)}{\epsilon_h} \Big] , \nonumber \\
p^{(\epsilon_y, \epsilon_h)} (y|x_i) &\propto \exp \Big\{ \frac{1}{\epsilon_y} \big[ \Delta(y, y_i)
 + \epsilon_h \log\sum_{h} \exp\big( \frac{ w^{T}\phi(x_i, y, h) }{ \epsilon_h} \big)  \big] \Big\}  , \nonumber\\
p^{(\epsilon_y, \epsilon_h)} (y, h|x_i) &= p^{\epsilon_h} (h|x_i, y) \cdot p^{(\epsilon_y, \epsilon_h)} (y|x_i). \nonumber
\end{align}
\begin{proof}
\begin{align}
\nabla_{w} \Big( \epsilon_h \log\sum_{h}\exp \big[ \frac{w^{T}\phi(x_i, {y}, h)}{\epsilon_h} \big] \Big)
&= \epsilon_h \frac{\sum_h \Big\{ \exp \Big[ \frac{w^{T}\phi(x_i, y, h)}{\epsilon_h} \Big]\cdot \Big[ \frac{\phi(x_i, y, h)}{\epsilon_h} \Big] \Big\} }
{\sum_h \exp \Big[ \frac{w^{T}\phi(x_i, y, h)}{\epsilon_h} \Big] } \nonumber \\
&= \sum_h \Big\{ \frac{ \exp \Big[ \frac{w^{T}\phi(x_i, y, h)}{\epsilon_h} \Big] } { \sum_h \exp \Big[ \frac{w^{T}\phi(x_i, y, h)}{\epsilon_h} \Big] } \cdot \phi(x_i, y, h) \Big\} \nonumber \\
&= \E_{p^{\epsilon_h}(h|x_i, y)}[\phi(x_i, y, h)] \label{Eqn2}
\end{align}
As a result, $ \nabla_{w} U_{i}^{-}(w;\epsilon_h) =  \E_{p^{\epsilon_h}(h|x_i, y_i)}[\phi(x_i, y_i, h)]$, and
\begin{align}
\nabla_{w} U_{i}^{+}(w; \epsilon_y, \epsilon_h)
&= \epsilon_y \frac{ \sum_{y} \Big\{ \exp \Big\{ \frac{1}{\epsilon_y} \Big[ \Delta(y_i, y) + \epsilon_h \log\sum_{h} \exp  \Big( \frac{w^{T}\phi(x_i, y, h)}{\epsilon_h} \Big) \Big] \Big\} \cdot \frac{1}{\epsilon_y} \cdot
\nabla_{w} \Big( \epsilon_h \log\sum_{h}\exp \big[ \frac{w^{T}\phi(x_i, {y}, h)}{\epsilon_h} \big] \Big) \Big\} }
{ \sum_{y} \exp \Big\{ \frac{1}{\epsilon_y} \Big[ \Delta(y_i, y) + \epsilon_h \log\sum_{h} \exp  \Big( \frac{w^{T}\phi(x_i, y, h)}{\epsilon_h} \Big) \Big] \Big\} }\nonumber \\
& \ \ \text{Subinstitute the gradient result (\ref{Eqn2}}), \nonumber \\
&= \frac { \sum_{y} \Big\{ \exp \Big\{ \frac{1}{\epsilon_y} \Big[ \Delta(y_i, y) + \epsilon_h \log\sum_{h} \exp  \Big( \frac{w^{T}\phi(x_i, y, h)}{\epsilon_h} \Big) \Big] \Big\} \cdot
\E_{p^{\epsilon_h}(h|x_i, y)}[\phi(x_i, y, h)] \Big\} }
{ \sum_{y} \exp \Big\{ \frac{1}{\epsilon_y} \Big[ \Delta(y_i, y) + \epsilon_h \log\sum_{h} \exp  \Big( \frac{w^{T}\phi(x_i, y, h)}{\epsilon_h} \Big) \Big] \Big\} }\nonumber \\
& =\E_{p^{(\epsilon_y, \epsilon_h)} (y|x_i)} \E_{p^{\epsilon_h}(h|x_i, y)} [\phi(x_i, y, h)] \nonumber \\
& = \E_{p^{(\epsilon_y, \epsilon_h)} (y, h|x_i)} [\phi(x_i, y, h)] 
\end{align}
which completes the proof.
\end{proof}
\end{lemma2}

\begin{figure*}[t!]
\vskip 0.15in
\begin{center}
\centerline{\includegraphics[width=6cm]{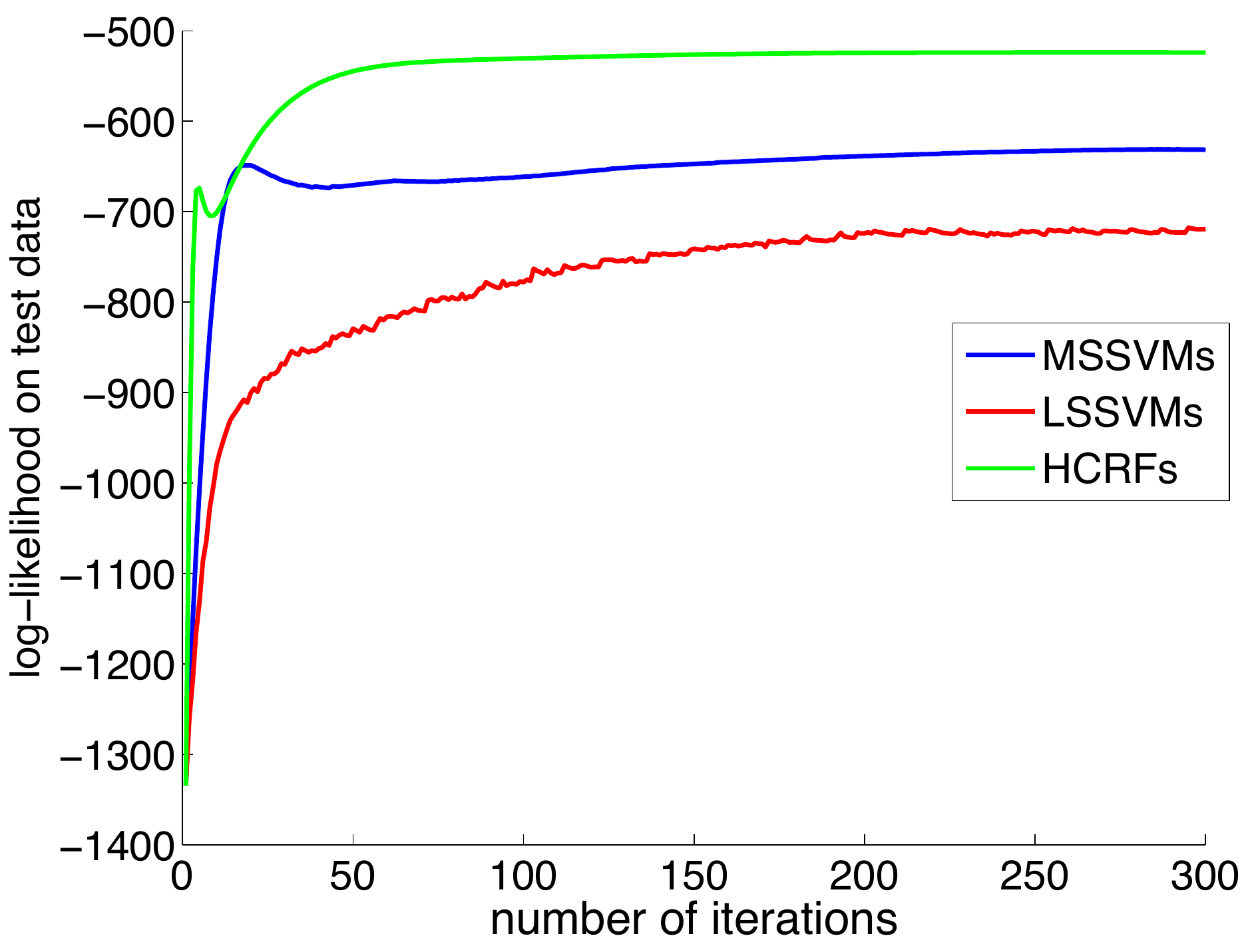}}
\caption{ The test log-likelihood of MSSVM, LSSVM and HCRF using SGD when 20 training and 100 test instances are sampled from 40-node hidden chain MRF (same setting as Table 2 in main paper).}
\end{center}
\vskip -0.2in
\label{fig:likelihood_iters}
\end{figure*}
\section*{Likelihood vs. Prediction Accuracy}
In our main paper, we demonstrate that our proposed MSSVM consistently outperforms HCRF on prediction accuracy.
However, it is worth noting that the HCRF model always achieves higher test likelihood than the MSSVM and LSSVM 
on our simulated data set.  As an example, Figure~5 shows the test log-likelikelihood 
across the different methods on these data.
This should not be surprising, since the HCRF model directly optimizes the likelihood objective, and (in this
case) the model class being optimized is correct (i.e., the data were drawn from a true model with the same structure).
However, higher likelihood does not necessarily imply that the HCRF will have better predictions on the
target variables. As was illustrated in the main paper (see details in Section 7.1, Training Sample Size), 
explicitly minimizing the empirical loss can lead to better predictions in situations with high dimensional 
model parameters and relatively few training instances.

\end{document}